\documentclass[conference]{IEEEtran}
\usepackage{times}
\usepackage{epsfig}
\usepackage{graphicx}
\usepackage{amsmath}
\usepackage{subfigure}
\usepackage{algorithm}
\usepackage{algorithmic}
\usepackage{amssymb}
\usepackage{multicol}
\usepackage{multirow}
\usepackage{enumitem}
\usepackage{stfloats}
\usepackage{amsthm}
\usepackage{cite}
\usepackage{color}
\usepackage{relsize}
\usepackage{booktabs}
\usepackage{mathtools}
\usepackage{cuted}
\usepackage{tikz}
\usetikzlibrary{shapes, arrows, calc, positioning,matrix}
\tikzset{
data/.style={circle, draw, text centered, minimum height=3em ,minimum width = .5em, inner sep = 2pt},
empty/.style={circle, text centered, minimum height=3em ,minimum width = .5em, inner sep = 2pt},
}
\usepackage{environ}
\makeatletter
\newsavebox{\measure@tikzpicture}
\NewEnviron{scaletikzpicturetowidth}[1]{%
  \def\tikz@width{#1}%
  \begin{lrbox}{\measure@tikzpicture}%
  \BODY
  \end{lrbox}%
  \pgfmathparse{#1/\wd\measure@tikzpicture}%
  \BODY
}
\makeatother

\newcommand{\normalpdf}[3]{\mathcal{N}\left(#1 \middle\vert#2 , #3\right)}

\theoremstyle{plain}
\newtheorem{theorem}{Theorem}[section]

\theoremstyle{definition}
\newtheorem{definition}[theorem]{Definition}
\theoremstyle{remark}

\usepackage{bm}
\DeclareMathAlphabet\mathbfcal{OMS}{cmsy}{b}{n}
\newcommand{\ten}[1]{\mathbfcal{#1}} 
\newcommand{\mat}[1]{\mathbf{#1}}

\usepackage[breaklinks=true,bookmarks=false]{hyperref}

\begin{document}

\title{Variational Bayesian Inference for Robust Streaming Tensor Factorization and Completion}
\author{\IEEEauthorblockN{Cole Hawkins}
\IEEEauthorblockA{Department of Mathematics, UC Santa Barbara \\
colehawkins@math.ucsb.edu}
\and
\IEEEauthorblockN{Zheng Zhang}
\IEEEauthorblockA{Department of ECE, UC Santa Barbara\\
zhengzhang@ece.ucsb.edu}
}

\maketitle

\begin{abstract}
Streaming tensor factorization is a powerful tool for processing high-volume and multi-way temporal  data in Internet networks, recommender systems and image/video data analysis. Existing streaming tensor factorization algorithms rely on least-squares data fitting and they do not possess a mechanism for tensor rank determination. This leaves them susceptible to outliers and vulnerable to over-fitting. This paper presents a Bayesian robust streaming tensor factorization model to identify sparse outliers, automatically determine the underlying tensor rank and accurately fit low-rank structure. We implement our model in Matlab and compare it with existing algorithms on tensor datasets generated from dynamic MRI and Internet traffic.
\end{abstract}
\section{Introduction}

Multi-way data arrays (i.e., tensors) are collected in various application domains including recommender systems~\cite{karatzoglou2010multiverse}, computer vision~\cite{liu2013tensor}, and uncertainty quantification~\cite{zhang2017big}. How to process, analyze and utilize such high-volume tensor data is a fundamental problem in machine learning and signal processing~\cite{KoldaBader}. Effective numerical techniques, such as CANDECOMP/PARAFAC (CP)~\cite{carroll1970analysis,harshman1970foundations} factorizations, have been proposed to compress full tensors and to obtain low-rank representations. Extensive optimization and statistical techniques have been developed to obtain low-rank factors and to predict the full tensor from an incomplete (and possibly noisy) multi-way data array~\cite{zhou2015bayesian,cpOptCompletion,Riemannian}.

Streaming tensors appear sequentially in the time domain. Incorporating temporal relationships in tensor data analysis can give significant advantages in anomaly detection~\cite{fanaee2016tensor}, discussion tracking~\cite{bader2008discussion} and context-aware recommender systems~\cite{tensor_recommender_systems}. In the past decade, streaming tensor factorization has been studied under several low-rank tensor models, such as the Tucker model in \cite{sun2006beyond} and the CP decomposition in \cite{shaden,online-cp,online-SGD,kasai2016online}. Most approaches use similar objective functions, but differ in choosing specific numerical optimization solvers. All existing streaming tensor factorizations assume a fixed rank, and no existing techniques can capture the sparse outliers in a streaming tensor. Several low-rank plus sparse techniques have been proposed for non-streaming tensor data~\cite{wright2009robust,zhao2015bayesian,zhao2016bayesian}.

{\bf Paper Contributions.} This paper proposes a method for the {\em robust factorization and completion} of streaming tensors. We model the whole temporal tensor as the sum of a low-rank streaming tensor and a time-varying sparse component. In order to capture these two different components, we present a Bayesian statistical model to enforce low rank and sparsity. The posterior probability density function (PDF) of the hidden factors is then computed by the variational Bayesian method~\cite{winn2005variational}. Our work can can be regarded as an extension of~\cite{babacan2012sparse,zhao2015bayesian,zhao2016bayesian} to streaming tensors with sparse outliers.

\section{Preliminaries and Notations}


We use a bold lowercase letter (e.g., $\mat{a}$) to represent a vector, a bold uppercase letter (e.g., $\mat{A}$) to represent a matrix, and a bold calligraphic letter (e.g., $\ten{A}$) to represent a tensor. An order-$N$ tensor is an $N$-way data array $\ten{A}\in \mathbb{R}^{I_1 \times I_2 \times \dots \times I_N}$, where $I_k$ is the size of mode $k$. Given integer $i_k \in [1, I_k] $ for $k=1, \ldots, N$, an entry of the tensor $\ten{A}$ is denoted as $a_{i_1,\cdots, i_N}$.

\begin{definition}
Let $\ten{A}$ and $\ten{B}$ be two tensors of the same dimensions. Their {\bf inner product} is defined as
\begin{equation}
\langle \ten{A},\ten{B} \rangle = \sum_{i_1=1}^{I_1} \dots \sum_{i_N=1}^{I_N}a_{i_1,\dots,i_N}b_{i_1,\dots,i_N}.\nonumber
\end{equation}
\end{definition}
The {\bf Frobenius norm} of tensor $\ten{A}$ is further defined as
\begin{equation}
\label{eq: tensor norm}
||\ten{A}||_{\rm F} = \langle \ten{A},\ten{A} \rangle^{1/2}.
\end{equation}

\begin{definition}
An $N$-way tensor $\ten{T} \in \mathbb{R}^{I_1\times \cdots \times I_N}$ is \textbf{rank-1} if it can be written as a single outer product of $N$ vectors
\begin{equation}
\ten{T} =\mat{a}^{(1)} \circ \dots \circ \mat{a}^{(N)}, \; \text{with} \; \mat{a}^{(k)} \in\mathbb{R}^{I_k} \; \text{for}\; k=1,\cdots, N. \nonumber
\end{equation}
\end{definition}

\begin{definition}\label{Def: CP}
The {\bf CP factorization}~\cite{carroll1970analysis,harshman1970foundations} expresses an $N$-way tensor $\ten{A}$ as the sum of multiple rank-1 tensors:
\begin{equation}
\ten{A} = \sum_{r=1}^Rs_r \mat{a}_r^{(1)} \circ \dots \circ \mat{a}_r^{(N)}, \; \text{with} \; \mat{a}_r^{(k)} \in\mathbb{R}^{I_k}.
\end{equation}
The minimal integer $R$ that ensures the equality is called the {\bf CP rank} of $\ten{A}$. 
We can also express the CP factorization as
\[
\ten{A} = \sum_{r=1}^R s_r\mat{a}_r^{(1)} \circ \dots \circ \mat{a}_r^{(N)} = [\![ \mat{A}^{(1)},\dots,\mat{A}^{(N)};\mat{s}]\!],
\]
where the mode-$k$ factors form the columns of matrix $\mat{A}^{(k)}$. 
\end{definition}
It is convenient to express $\mat{A}^{(k)}$ both column-wise and row-wise, so we include two means of expressing a factor matrix
\[
\mat{A}^{(k)} = [\mat{a}_1^{(k)},\dots ,\mat{a}_R^{(k)}]=[\hat{\mat{a}}_1^{(k)};\dots ;\hat{\mat{a}}_{I_k}^{(k)}] \in \mathbb{R}^{I_k\times R}.
\]
Here $\mat{a}_j^{(k)}$ and $\hat{\mat{a}}_{i_k}^{(k)}$ denote the $j$-th column and $i_k$-th row of $\mat{A}^{(k)}$, respectively. 


\begin{definition}
The {\bf generalized inner product} of $N$ vectors of the same dimension $I$ is defined as
\[
\langle \mat{a}^{(1)},\dots,\mat{a}^{(N)} \rangle = \sum\limits_{i=1}^I\prod\limits_{k=1}^N a^{(k)}_i.
\]
\end{definition}

With a generalized inner product, the entries of a low-rank tensor $\ten{A}$ as in Definition \ref{Def: CP} can be written as:
\[
a_{i_1,\dots,i_N} = \langle \hat{\mat{a}}_{i_1}^{(1)},\dots,\hat{\mat{a}}_{i_N}^{(N)} \rangle.
\]

The {\bf Khatri-Rao product} of two matrices $\mat{A} \in \mathbb{R}^{I\times R}$ and $\mat{B} \in \mathbb{R}^{J\times R}$ is the columnwise Kronecker product:
\[
\mat{A}\odot\mat{B}= [\mat{a}_1 \otimes \mat{b}_1, \ldots, \mat{a}_R \otimes \mat{b}_R] \in \mathbb{R}^{IJ\times R}.
\]
We will use the product notation to denote the Khatri-Rao product of $N$ matrices in reverse order:
\[
\bigodot_n \mat{A}^{(n)} = \mat{A}^{(N)}\odot \mat{A}^{(N-1)}\odot\cdots\odot\mat{A}^{(1)}.
\]
If we exclude the $k$-th factor matrix, the Khatri-Rao product can be written as
\[
\bigodot_{n\neq k} \mat{A}^{(n)} = \mat{A}^{(N)}\odot\cdots\odot \mat{A}^{(k+1)}\odot\mat{A}^{(k-1)}\odot\mat{A}^{(1)}.
\]

\section{Review of Streaming Tensor Factorization}

Let $\{\ten{X}_t\}$ be a temporal sequence of $N$-way tensors, where $t \in \mathbb{N}$ is the time index and the tensor $\ten{X}_t$ of size $I_1\times \dots \times I_N$ is a slice of this multi-way stream. Streaming tensor factorization aims to extract the latent CP factors evolving with time.

The standard streaming tensor factorization is~\cite{online-SGD}:
\begin{equation}
\label{eq: basic streaming}
\min_{\{\mat{A}^{(k)}\}_{k=1}^{N+1}} \sum_{t=i}^T \mu^{T-t} \|\ten{X}_t - [\![ \mat{A}^{(1)},\dots,\mat{A}^{(N)};\hat{\mat{a}}^{(N+1)}_{t-i+1}]\!]\|_F^2.
\end{equation}
The parameter $\mu\in(0,1)$ is a forgetting factor that controls the weight of the past data; $\{\mat{A}^{(i)}\}$ are the discovered CP factors. Please note that $\hat{\mat{a}}^{(N+1)}_{t-i+1}$ denotes one row of the temporal factor matrix $\mat{A}^{(N+1)}$. The sliding window size $T-i+1$ can be specified by the user.

In many applications, only partial data $\ten{X}_{t,\Omega_t}$ is observed at each time point. Here $\Omega_t$ denotes the index set of the partially observed entries. For a general $N$-way tensor $\ten{X}$ and a sampling set $\Omega$, we have
\begin{equation}
\ten{X} _{\Omega }=\left\{\begin{matrix}
x_{i_1,\cdots,i_N}& \; {\rm if} \; {(i_1,i_2,\cdots,i_N)} \in \Omega\\ 
0 & \; {\rm otherwise}. \;\;\;\;\;\;\;\;\;\;
\end{matrix}\right. \nonumber
\end{equation}
In such cases, the underlying hidden factors can be computed by solving the following {\it streaming tensor completion} problem:
\begin{equation}
\label{eq: streaming completion}
\min_{\{\mat{A}^{(k)}\}_{k=1}^{N+1}} \sum_{t=i}^T \mu^{T-t} \|\left(\ten{X}_t - [\![ \mat{A}^{(1)},\dots,\mat{A}^{(N)};\hat{\mat{a}}^{(N+1)}_{t-i+1} ]\!]\right)_{\Omega_t} \|_{\rm F}^2.
\end{equation}

Existing streaming factorization and completion frameworks~\cite{online-cp,online-SGD,kasai2016online} solve \eqref{eq: basic streaming} and \eqref{eq: streaming completion}  as follows: at each time step one updates the $N$ non-temporal factor matrices $\mat{A}^{(j)}\in\mathbb{R}^{I_j\times R}$ and  $\{\hat{\mat{a}}^{(N+1)}_{t-i+1}\}$.
By fixing the past time factors, these approaches provide an efficient updating scheme to solve the above non-convex problems.

\section{Bayesian Model for Robust Streaming Tensor Factorization \& Completion}

In this section, we present a Bayesian method for the robust factorization and completion of streaming tensors $\{ \ten{X}_t\}$.

\subsection{An Optimization Perspective}
In order to simultaneously capture the sparse outliers and the underlying low-rank structure of a streaming tensor, we assume that each tensor slice $\ten{X}_t$ can be fit by
\begin{equation}\label{eq:slice L+S}
\ten{X}_t = \tilde{\ten{X}}_t+\ten{S}_t+\ten{E}_t.
\end{equation}
Here $\tilde{\ten{X}}_t$ is low-rank, $\ten{S}_t$ contains sparse outliers, and $\ten{E}_t$ denotes dense noise with small magnitudes. Assume that each slice $\ten{X}_t$ is partially observed according to a sampling index set $\Omega_t$. Based on the partial observations $\{ \ten{X}_{t,\Omega_t}\}$, we will find reasonable low-rank factors for $\{\tilde{\ten{X}}_t\}$ in the specified time window $t \in [T-i+1, T]$ as well as the sparse component $\ten{S}_t$. This problem simplifies to robust streaming tensor factorization if $\Omega_t$ includes all possible indices.

In order to enforce the low-rank property of $\tilde{\ten{X}}_t $, we assume the following CP representation for $t\in [T-i+1, T]$:
\begin{equation}
\tilde{\ten{X}}_t=[\![ \mat{A}^{(1)},\dots,\mat{A}^{(N)};\hat{\mat{a}}^{(N+1)}_{t-i+1}]\!]. \nonumber
\end{equation}
The sparsity of $\ten{S}_t$ can be enforced by adding an $L_1$ regularizer and modifying \eqref{eq: streaming completion} as follows: 

\begin{align}
\label{eq: sparse completion streaming} 
 \nonumber\min_{\{\mat{A}^{(j)}\},\ten{S}_{\Omega_T}} \ &\sum_{t=i}^{T-1} \mu^{T-t} \|\left(\tilde{\ten{D}}_t  - [\![\mat{A}^{(1)},\dots,\mat{A}^{(N)};\hat{\mat{a}}^{(N+1)}_{t-i+1} ]\!]\right)_{\Omega_t}\|_{\rm F}^2\\ 
 \nonumber +&\|\ten{Y}_{\Omega_T}-\ten{S}_{\Omega_T}-\left([\![\mat{A}^{(1)},\dots,\mat{A}^{(N)};\hat{\mat{a}}^{(N+1)}_{T-i+1} ]\!]\right)_{\Omega_T}\|_{\rm F}^2\\
 +&\alpha\|\ten{S}_{\Omega_T}\|_1.
\end{align}
Here $\ten{Y}_{\Omega_T}=\ten{X}_{T,\Omega_T}$ is the observation of current slice, $\ten{S}_{\Omega_T}$ is its outliers, and $\{\tilde{\ten{D}}_{t,\Omega_t}\}_{t=i}^{T-1}$ are the observed past slices {\em with their sparse errors removed}. Based on the results of of all previous slices, $\tilde{\ten{D}}_t$ is obtained as $\tilde{\ten{D}}_t=\ten{X}_t -\ten{S}_t$.

It is challenging to determine the rank $R$ in (\ref{eq: sparse completion streaming}). It is also non-trivial to select a proper regularization parameter $\alpha$. In order to fix these issues, we develop a Bayesian model.

\subsection{Probabilistic Model for (\ref{eq:slice L+S}) }
{\bf Likelihood:} We first define a likelihood function for the data  $\ten{Y}_{\Omega_T}$ and $\{\tilde{\ten{D}}_{t,\Omega_t}\}_{t=i}^{T-1}$ based on (\ref{eq:slice L+S}) and \eqref{eq: sparse completion streaming}. We discount the past observations outside of the time window, and use the forgetting factor $\mu< 1$ to exponentially weight the variance terms of past observations. This permits long-past observations to deviate significantly from the current CP factors with little impact on the current CP factors. Therefore, at time point $t$, we assume that the Gaussian noise has a 0 mean and variance $(\mu^{T-t} \tau)^{-1}$. This leads to the likelihood function in (\ref{eq:observationModel}). In this likelihood function, $\tau$ specifies the noise precision, $\hat{\mat{a}}_{i_n}^{(n)}$ denotes the $i_n$-th row of $\mat{A}^{(n)}$, and $\ten{S}_{\Omega_T}$ only has values corresponding to observed locations. In order to infer the unknown factors, we also specify their prior distributions. 
\begin{figure*}[t]
\begin{align}
\label{eq:observationModel}
p\left(\ten{Y}_{\Omega_T},\{\tilde{\ten{D}}_{t,\Omega_t}\}\middle\vert\{\mat{A}^{(n)}\}_{n=1}^{N+1},\ten{S}_{\Omega_T},\tau\right) = & \prod_{(i_1,\dots,i_n)\in \Omega_T}
\mathcal{N}\left(\ten{Y}_{i_1 \ldots i_N} \middle\vert \left\langle\hat{\mat{a}}^{(1)}_{i_1},\cdots,\hat{\mat{a}}^{(N)}_{i_N},\hat{\mat{a}}^{(N+1)}_{T-i+1}\right\rangle + \mathcal{S}_{i_1 \ldots i_N}, \tau^{-1}\right) \times \nonumber \\
&\prod_{t=i}^{T-1}\prod_{(i_1,\dots,i_n)\in\Omega_t}\mathcal{N}\left({\tilde{\ten{D}}}_{t,{i_1 \ldots i_N}} \middle\vert \left\langle\hat{\mat{a}}^{(1)}_{i_1},\cdots,\hat{\mat{a}}^{(N)}_{i_N},\hat{\mat{a}}^{(N+1)}_{t-i+1},\right\rangle, (\tau\mu^{T-t})^{-1}\right).
\end{align}
\normalsize
\end{figure*}

{\bf Prior Distribution of $\{ \mat{A}^{(n)}\}$:} We assume that each row of $\mat{A}^{(n)}$ obeys a Gaussian distribution and that different rows are independent to each other. Similar to~\cite{zhao2015bayesian}, we define the prior distribution of each factor matrix as
\begin{equation}
\label{eq:priorA}
p\big(\mat{A}^{(n)}\big\vert \boldsymbol\lambda \big) = \prod_{i_n=1}^{I_n} \mathcal{N}\big(\hat{\mat{a}}_{i_n}^{(n)} \big\vert \mathbf{0}, \boldsymbol\Lambda^{-1} \big), \, \forall n\in [1,N+1] 
\end{equation}
where $\boldsymbol\Lambda=\text{diag}(\boldsymbol\lambda)\in\mathbb{R}^{R\times R}$ denotes the precision matrix. All factor matrices share the same covariance matrix. The $r$-th column of all factor matrices share the same precision parameter $\lambda_r$, and a large $\lambda_r$ will make the $r$-th rank-1 term more likely to have a small magnitude. Therefore, the hyper parameters $\boldsymbol\lambda \in \mathbb{R}^R$ can tune the rank of our CP model.

{\bf Prior Distribution of $\ten{S}_{{\Omega}_T}$:} We also place a Gaussian prior distribution over the component $\ten{S}_{ \Omega_T}$:
\begin{equation}
\label{eq:priorS}
p(\ten{S}_{\Omega_T}\vert\boldsymbol\gamma) = \prod_{(i_1,\ldots,i_N) \in \Omega_T}\mathcal{N}(\mathcal{S}_{i_1 \ldots i_N}\vert 0, \gamma_{i_1 \ldots i_N}^{-1}),
\end{equation}
where $\boldsymbol{\gamma}$ is the sparsity precision matrix. If $\gamma_{i_1 \ldots i_N}$ is very large, then the associated element in $\ten{S}_{\Omega_T}$ is likely to have a very small magnitude. By controlling the value of $\gamma_{i_1 \ldots i_N}^{-1}$, we can control the sparsity of $\ten{S}_{\Omega_T}$. 

\begin{figure*}
\begin{equation}\label{eq:posterior density}
p\left(\Theta\middle\vert\ten{Y}_{\Omega_T},\{\tilde{\ten{D}}_{t,\Omega_t}\}\right) =
\frac{p\left(\ten{Y}_{\Omega_T},\{\tilde{\ten{D}}_{t,\Omega_t}\}\middle\vert\ \{\mat{A}^{(n)}\}_{n=1}^{N+1},\ten{S}_{\Omega_T},\tau \right)
\left\{
{\prod\limits_{n=1}^{(N+1)}p\big(\mat{A}^{(n)}}\big\vert \boldsymbol\lambda \big) \right\} p(\boldsymbol\lambda)
p(\ten{S}_{\Omega_T}\vert\boldsymbol\gamma)p(\boldsymbol\gamma)p(\tau)}{p(\ten{Y}_{\Omega_T},\{\tilde{\ten{D}}_{t,\Omega_t}\})}.
\end{equation}
\end{figure*}

\subsection{Prior Distribution of Hyper Parameters}
\label{subsec: hyper}

We still have to specify three groups of hyper parameters: $\tau$ controlling the noise term, $\boldsymbol \lambda$ controlling the CP rank, and $\{ \gamma_{i_1 \ldots i_N}\}$ controlling the sparsity of $\ten{S}_{\Omega_T}$. We treat them as random variables and assign them Gamma prior distributions:
\begin{equation}
\label{eq:prior_hyper}
\begin{split}
p(\tau) &= \text{Ga}(\tau\:\vert \: a_0^\tau, b_0^\tau),\;\;
p(\boldsymbol\lambda) = \prod_{r=1}^{R}\text{Ga}(\lambda_r \vert c_0, d_0),\\
p(\boldsymbol\gamma) &= \prod_{(i_1,\ldots,i_N)\in\Omega_T}\text{Ga}(\gamma_{i_1 \ldots i_N }\vert a_0^\gamma, b_0^\gamma).
\end{split}
\end{equation}

The Gamma distribution provides a good model for our hyper parameters due to its non-negativity and its long tail. The mean value and variance of the above Gamma distribution are $a/b$ and $a/b^2$, respectively. 

These hyper parameters control $\{\mat{A}^{(n)}\}$ and $\ten{S}$. For instance, the noise term tends to have a very small magnitude if $\tau$ has a large mean value and a small variance; if $\lambda_r$ has a large mean value, then the $r$-th rank-1 term in the CP factorization tends to vanish, leading to rank reduction.

\subsection{Posterior Distribution of Model Parameters}
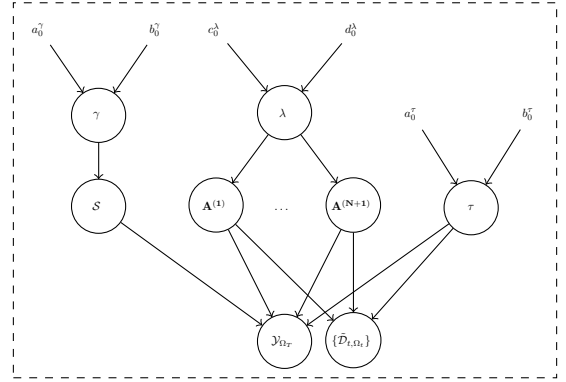
\begin{figure}[t]
\begin{center}
\begin{tikzpicture}[scale=0.45, every node/.style={scale=0.45},align = flush center,
data/.style={circle, draw, minimum size=1.6cm}
]
\matrix [
    matrix of nodes,
    column sep = 0.5em,
    row sep = 1.3em, 
    draw, dashed,
    nodes = {solid},
    ] (m)
{  
|[empty]|$a_0^\gamma$ &  |[empty]| & |[empty]|$b_0^\gamma$ & |[empty]|$c_0^\lambda$ & |[empty]| & |[empty]|$d_0^\lambda$ & |[empty]| & |[empty]|& |[empty]| \\
|[empty]| &  |[data]|$\gamma$ & |[empty]| & |[empty]| & |[data]|$\lambda$ & |[empty]| & |[empty]|$a_0^\tau$ & |[empty]|& |[empty]|$b_0^\tau$ \\
|[empty]| &  |[data]|$\mathcal{S}$ & |[empty]| & |[data]| $\bf{A}^{(1)}$ & |[empty]|$\dots$ & |[data]|$\bf{A}^{(N+1)}$ & |[empty]| & |[data]|$\tau$& |[empty]|\\
[1.6em]
|[empty]| &  |[empty]| & |[empty]| & |[empty]| & |[data]|$\mathcal{Y}_{\Omega_T}$ & |[data]|$\{\tilde{\mathcal{D}}_{t,\Omega_t}\}$ & |[empty]| & |[empty]|& |[empty]|\\[.7em]
};
\draw[->](m-1-1) to (m-2-2);
\draw[->](m-1-3) to (m-2-2);
\draw[->](m-2-2) to (m-3-2);
\draw[->](m-3-2) to (m-4-5);

\draw[->](m-1-4) to (m-2-5);
\draw[->](m-1-6) to (m-2-5);
\draw[->](m-2-5) to (m-3-4);
\draw[->](m-2-5) to (m-3-6);
\draw[->](m-3-4) to (m-4-5);
\draw[->](m-3-4) to (m-4-6);
\draw[->](m-3-6) to (m-4-5);
\draw[->](m-3-6) to (m-4-6);

\draw[->](m-2-9) to (m-3-8);
\draw[->](m-2-7) to (m-3-8);
\draw[->](m-3-8) to (m-4-6);
\draw[->](m-3-8) to (m-4-5);
\end{tikzpicture}
\caption{The probabilistic graphical model for our Bayesian robust streaming tensor completion.}
\label{fig:graphical model}
\end{center}
\end{figure}

Now we can present a graphical model describing our Bayesian formulation in Fig.~\ref{fig:graphical model}. Our goal is to infer all hidden parameters based on partially observed data. For convenience, we denote all unknown hidden parameters in a compact form:
\begin{equation}
\Theta=\left \{ \{\mat{A}^{(n)}\}_{n=1}^{N+1},\ten{S}_{\Omega_T},\tau,\boldsymbol\lambda,\boldsymbol\gamma \right\}. \nonumber
\end{equation}
With the likelihood function \eqref{eq:observationModel}, prior distribution for low-rank factors and sparse components in \eqref{eq:priorA} and \eqref{eq:priorS}, and prior distribution of the hyper-parameters in \eqref{eq:prior_hyper}, we can obtain the posterior distribution in 
(\ref{eq:posterior density}) using Bayes theorem.


The main challenge is how to estimate the resulting posterior distribution (\ref{eq:posterior density}). We address this issue in Section~\ref{sec:update}.

\section{Variational Bayesian Solver For Model Parameter Estimation}
\label{sec:update}
It is hard to obtain the exact posterior distribution (\ref{eq:posterior density}) because the marginal density $p(\ten{Y}_{\Omega_T},\{\tilde{\ten{D}}_{t,\Omega_t}\})$ is unknown and is expensive to compute. Therefore, we employ variational Bayesian inference~\cite{winn2005variational} to obtain a closed-form approximation of the posterior density (\ref{eq:posterior density}). The variational Bayesian method was previously employed for matrix completion~\cite{babacan2012sparse} and non-streaming tensor completion~\cite{zhao2015bayesian, zhao2016bayesian}, and it is a popular inference technique in many domains. We use a similar procedure to ~\cite{babacan2012sparse, zhao2015bayesian} to derive our iteration steps, but the details are quite different since we solve a streaming problem and we approximate an entirely different posterior distribution. 

\begin{algorithm}[t]
    \caption{Variational Bayesian Updating Scheme for Streaming Tensor Completion}
    \label{algoflow}
    \begin{algorithmic}
    \WHILE{Not Converged}
\STATE Update the variance matrices via Equations (\ref{eq:updatevar},\ref{eq:updatetimevariance})
\STATE Update the factor matrices by Equations (\ref{eq:update non time matrix},\ref{eq:update time matrix}, \ref{eq:updatetimevariance2})
\STATE Update the rank prior $\boldsymbol \lambda$ by Equation (\ref{eq:update lambda})
\STATE Update the sparse term $\ten{S}_{\Omega_T}$ by Equation (\ref{eq:update S})
\STATE Update the sparsity prior $\boldsymbol  \gamma$ by Equation (\ref{eq:update gamma})
\STATE Update the precision $\tau$ by Equation (\ref{eq:update tau})
\ENDWHILE

    \end{algorithmic}
\end{algorithm}

Due to the complexity of our analysis and the page space limitation, we provide only the main idea and key results of our solver. The algorithm flow is summarized in Alg.~\ref{algoflow}, and the complete derivations are available in~\cite{hawkins2018arxiv}.

\subsection{Variational Bayesian}

We intend to find a distribution $q(\Theta)$ that approximates the true posterior distribution $p(\Theta\vert \ten{Y}_{\Omega_T},\{\tilde{\ten{D}}_{t,\Omega_t}\})$ by minimizing the following KL divergence:
\begin{equation}\label{eq:VBKL}
\begin{split}
\small
\text{KL}\big(q(\Theta)\big\vert\big\vert p(\Theta\vert \ten{Y}_{\Omega_T},\{\tilde{\ten{D}}_{t,\Omega_t}\})\big)= \ln p(\ten{Y}_{\Omega_T},\{\tilde{\ten{D}}_{t,\Omega_t}\}) &- \mathcal{L}(q), \\
 \mbox{where} \; \mathcal{L}(q) = \int q(\Theta)\ln \left(\frac{p(\ten{Y}_{\Omega_T}, \{\tilde{\ten{D}}_{t,\Omega_t}\},\Theta)}{q(\Theta)}\right)&d\Theta.
\end{split}
\end{equation}

The quantity $\ln p(\ten{Y}_{\Omega_T},\{\tilde{\ten{D}}_{t,\Omega_t}\})$ denotes model evidence and is a constant. Therefore, minimizing the KL divergence is equivalent to maximizing $\mathcal{L}(q)$. To do so we apply mean field variational approximation. That is, we assume that the posterior can be factorized as a product of the individual marginal distributions:
\begin{equation}
\label{eq:vbfactorization}
q\left(\Theta\right) =\left\{ \prod_{n=1}^{N+1} q\left(\mat{A}^{(n)}\right) \right\}q(\ten{S}_{\Omega_T}) q(\boldsymbol\lambda)q(\boldsymbol\gamma) q(\tau).
\end{equation}
This assumption allows us to maximize $\mathcal{L}(q)$ by applying the following update rule
\begin{equation}\label{eq: param update}
\ln q(\Theta_i) \propto \max_{\Theta_i}\mathbb{E}_{\Theta_{j\neq i}}\ln (p(\ten{Y}_{\Omega_T},\{\tilde{\ten{D}}_{t,\Omega_t}\},\Theta)),
\end{equation}
where the subscript $\Theta_{j\neq i}$ denotes the expectation with respect to all latent factors except $\Theta_i$. In the following we will provide the closed-form expressions of these alternating updates. 

\subsection{Factor Matrix Updates}

The posterior distribution of an individual factor matrix is
\begin{equation*}\label{eq:psoteriorA}
q\big(\mat{A}^{(n)}) = \prod_{i_n=1}^{I_n} \mathcal{N}\big(\hat{\mat{a}}^{(n)}_{i_n}\big\vert \bar{\mat{a}}_{i_n}^{(n)},\mat{V}_{i_n}^{(n)}\big).\end{equation*}
Therefore, we must update the posterior mean $\bar{\mat{a}}_{i_n}^{(n)}$ and co-variance $\mat{V}_{i_n}^{(n)}$ for each row of $\mat{A}^{(n)}$.

{\bf Update non-temporal factors.} {All non-time factors are updated by Equations (\ref{eq:updatevar}) and (\ref{eq:update non time matrix}) for $ n\in [1,\dots,N]$:

\begin{equation}\label{eq:updatevar}
\small
{\mat{V}}^{(n)}_{i_n} = \left(\mathbb{E}_q[\tau] \sum_{t=i}^T{\mu^{T-i}}\mathbb{E}_q\left[\mat{A}_{i_n}^{(\setminus n)T}\mat{A}_{i_n}^{(\setminus n)}\right]_{\Omega_t} +\mathbb{E}_q[\boldsymbol\Lambda]\right)^{-1},
\end{equation}
\begin{multline}\label{eq:update non time matrix}
\small
{\bar{\mat{a}}}^{(n)}_{i_n} = \mathbb{E}_q[\tau]\mat{V}^{(n)}_{i_n}\Bigg(\mathbb{E}_q\left[\mat{A}_{i_n}^{(\setminus n)T}\right]_{\Omega_T}\text{vec}\left( \ten{Y}_{\Omega_T}-\mathbb{E}_q[\ten{S}_{\Omega_T}]\right) \\
+\sum_{t=i}^{T-1} \mu^{T-t}\mathbb{E}_q\left[\mat{A}_{i_n}^{(\setminus n)T}\right]_{\Omega_t}\text{vec}\left(\tilde{\ten{D}}_{t,\Omega_t}\right) \Bigg).
\end{multline}
The notation $\mathbb{E}_q\left[\mat{A}_{i_n}^{(\setminus n)}\right]_{\Omega_t}$ represents a sampled expectation of the excluded Khatri-Rao product:
\[
\mathbb{E}_q\left[\mat{A}_{i_n}^{(\setminus n)}\right]_{\Omega_t} = \left(\mathbb{E}_q\left[\bigodot_{j\neq n} \mat{A}^{(j)}\right]\right)_{\mathbb{I}_{i_n}}.
\]
The matrix $\mat{A}_{i_n}^{(\setminus n)}$ is $\prod_{j\neq n}I_j\times R$ and the indicator function ${\mathbb{I}_{i_n}}$ samples the row $(i_1,\dots, i_{n-1},i_{n+1},\dots, i_{N+1})$ if the entry $(i_1,\dots, i_{n-1},i_n,i_{n+1},\dots,i_{N+1})$ is in $\Omega_t$ and sets the row to zero if not. The expression $\mathbb{E}_q[\cdot]$ denotes the posterior expectation with respect to all variables involved. 

{\bf Update temporal factors.} The temporal factor $\mat{A}^{N+1}$ requires a different update because the factors corresponding to different time slices do not interact with each other. For all time factors the variance is updated according to
\begin{align}\label{eq:updatetimevariance}
\small
\begin{split}
{\mat{V}}^{(N+1)}_{t-i+1} = \left(\mathbb{E}_q[\tau] {\mu^{T-t}}\mathbb{E}_q\left[\mat{A}_{t-i+1}^{(\setminus (N+1))T}\mat{A}_{t-i+1}^{(\setminus (N+1))}\right]_{\Omega_t} +\mathbb{E}_q[\mat{\Lambda}]\right)^{-1}.
\end{split}
\end{align}
The rows of the time factor matrix are updated differently depending on their corresponding time indices. Since we assume that past observations have had their sparse errors removed, the time factors of all past slices (so $t< T$) can be updated by
\begin{align}\label{eq:update time matrix}
\small
\begin{split}
{\bar{\mat{a}}}^{(N+1)}_{t-i+1} = \mathbb{E}_q[\tau]\mat{V}^{(N+1)}_{t-i+1}\left(\mu^{T-t}\mathbb{E}_q\left[\mat{A}_{t-i+1}^{(\setminus N+1)T}\right]_{\Omega_t}\text{vec}\left(\tilde{\ten{D}}_{t,\Omega_t}\right) \right).
\end{split}
\end{align}
The factors corresponding to time slice $T$ depend on the sparse errors in the current step. The update is therefore given by
\begin{align}\label{eq:updatetimevariance2}
\small
\begin{split}
{\bar{\mat{a}}}^{(N+1)}_{T-i+1} = \mathbb{E}_q[\tau]\mat{V}^{(N+1)}_{T-i+1}\left(\mathbb{E}_q\left[\mat{A}_{T-i+1}^{(\setminus N+1)T}\right]_{\Omega_T}\text{vec}\left(\ten{Y}_{\Omega_T}-\mathbb{E}_q\left[\ten{S}_{\Omega_T}\right]\right) \right).
\end{split}
\end{align}


\subsection{Posterior Distribution of Hyperparameters $\boldsymbol\lambda$}

The posteriors of the parameters $\lambda_r$ are independent Gamma distributions. Therefore the joint distribution takes the form
\[
q(\boldsymbol\lambda) = \prod_{r=1}^{R} \text{Ga}(\lambda_r \vert {c}_M^r, {d}_M^r )
\]
where $c_M^r$, $d_M^r$ denote the posterior parameters learned from the previous $M$ iterations. The updates to $\boldsymbol{\lambda}$ are given below.
\begin{equation}\label{eq:update lambda}
c_M^r = c_0+1 + \frac{1}{2}\sum_{n=1}^{N} I_n, \quad
d_M^r = d_0 + \frac{1}{2}\sum_{n=1}^{N+1} \mathbb{E}_q\left[{\mat{a}}^{(n)T}_{ r}{\mat{a}}^{(n)}_{ r}\right]
\end{equation}

The expectation of each rank-sparsity parameter can then be computed as
\[
\mathbb{E}_q[\boldsymbol\Lambda] = \text{diag}([c_M^1/d_M^1,\ldots,c_M^R/d_M^R]).
\]

\subsection{Posterior Distribution of Sparse tensor $\ten{S}$}
The posterior approximation of $\ten{S}_{\Omega_T}$ is given by
\begin{equation}
q(\ten{S}_{\Omega_T}) = \prod_{(i_1,\ldots,i_N)\in\Omega_T} \normalpdf{\mathcal{S}_{i_1 \ldots i_N} }{\bar{\mathcal{S}}_{i_1 \ldots i_N}}{\sigma^2_{i_1 \ldots i_N}},
\end{equation}
where the posterior parameters can be updated by
\begin{gather}\label{eq:update S}
\begin{aligned}
{\bar{\mathcal{S}}_{i_1 \ldots i_N}} =& \sigma^2_{i_1 \ldots i_N}\mathbb{E}_q[\tau]\Big(\mathcal{Y}_{i_1\ldots i_N}-\\
&\mathbb{E}_q\left[\left\langle\hat{\mat{a}}_{i_1}^{(1)},\ldots,\hat{\mat{a}}_{i_N}^{(N)};\hat{\mat{a}}_{T-i+1}^{(N+1)}\right\rangle \right] \Big) \\
\sigma^2_{i_1 \ldots i_N} =& (\mathbb{E}_q[\gamma_{i_1\ldots i_N}] + \mathbb{E}_q[\tau])^{-1}.
\end{aligned}
\end{gather}


\subsection{Posterior Distribution of Hyperparameters $\boldsymbol\gamma$}

The posterior of $\boldsymbol\gamma$ is also factorized into entry-wise independent distributions
\begin{equation}
q(\boldsymbol\gamma) = \prod_{(i_1,\ldots,i_N)\in\Omega_T}\text{Ga}(\gamma_{i_1 \ldots i_N} \vert {a}_M^{\gamma_{i_1 \ldots i_N}}, {b}_M^{\gamma_{i_1 \ldots i_N}} ),
\end{equation}
whose posterior parameters can be updated by
\begin{equation}\label{eq:update gamma}
{a}_M^{\gamma_{i_1 \ldots i_N}} = a_0^\gamma + \frac{1}{2}, \hspace{.04in}
{b}_M^{\gamma_{i_1 \ldots i_N}} = b_0^\gamma + \frac{1}{2}({\bar{\mathcal{S}}}^2_{i_1 \ldots i_N} + \sigma^2_{i_1 \ldots i_N}).
\end{equation}

\subsection{Posterior Distribution of Parameter $\tau$}
The posterior PDF of the noise precision is again a Gamma distribution. The posterior parameters can be updated by
\begin{gather}\label{eq:update tau}
\begin{aligned}
\small
{a}_M^\tau = &\frac{1}{2}|{\Omega_T}|+ a_0^\tau ,\\
{b}_M^\tau =  & \frac{1}{2}\mathbb{E}_q\left[\|\left(\ten{Y}-{\ten{S}}- [\![ {\mat{A}}^{(1)},\ldots,{\mat{A}}^{(N)}; \hat{{\mat{a}}}_{T-i+1}^{(N+1)} ]\!] \right)_{\Omega_T}\|_{\rm F}^2\right]\\
& +b_0^\tau.
\end{aligned}
\end{gather}

\section{Numerical Results}
Our algorithm is implemented in Matlab and is compared with several existing streaming tensor factorization and completion methods. These include Online-CP~\cite{online-cp}, Online-SGD~\cite{online-SGD} and OLSTEC \cite{kasai2016online}. Both Online-CP and OLSTEC solve essentially the same optimization problem, but Online-CP does not support incomplete tensors. Therefore, our algorithm is only compared with OLSTEC and Online-SGD for the completion task. Our Matlab codes to reproduce all figures and results can be downloaded from \url{https://github.com/colehawkins}. We provide more extensive numerical results, including results on surveillance video and automatic rank determination, in~\cite{hawkins2018arxiv}.

\subsection{Dynamic Cardiac MRI}

We consider a dynamic cardiac MRI dataset from \cite{sharif2007physiologically} and obtained via \url{https://statweb.stanford.edu/~candes/SURE/data.html}. Each slice of this streaming tensor dataset is a $128\times 128$ matrix. In clinical applications, it is highly desirable to reduce the number of MRI scans. Therefore, we are interested in using streaming tensor completion to reconstruct the whole sequence of medical images based on a few sampled entries. 

In all methods we set the maximum rank to $15$. For our algorithm we set the forgetting factor to $\mu = 0.98$ and the the sliding window size to $20$. In OLSTEC we set the forgetting factor to the suggested default of $0.7$ and the sliding window size to $20$. The available implementation of Online-SGD does not admit a sliding window, but instead computes with the full (non-streamed) tensor. While this may limit its ability to work with large streamed data in practice, we include it in comparison for completeness. With $15\%$ random samples, the reconstruction results are shown in Fig.~\ref{fig:mri reconstruction}. The ability of our model to capture both small-magnitude measurement noise and sparse large-magnitude deviations renders it more effective than OLSTEC and Online-SGD for this dynamic MRI reconstruction task.

\begin{figure}[t]
\begin{center}
\includegraphics[width=.45\textwidth,height=5cm]{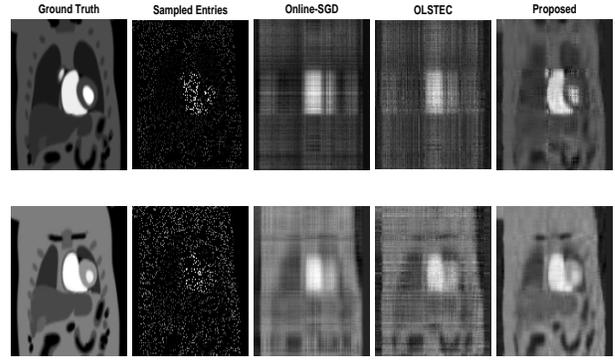}
\caption{MRI reconstruction via streaming tensor completion.}
\label{fig:mri reconstruction}
\end{center}
\end{figure}

\subsection{Network Traffic}
Our next example is the Abilene network traffic dataset~\cite{lakhina2004structural}. This dataset consists of aggregate Internet traffic between 11 nodes, measured at five-minute intervals. On this dataset we test our algorithm for both reconstruction and completion. The goal is to identify abnormally evolving network traffic patterns between nodes. If one captures the underlying low-rank structure, one can identify anomalies for further inspection. Anomalies can range from malicious distributed denial of service (DDoS) attacks to non-threatening network traffic spikes related to online entertainment releases. In order to classify abnormal behavior one must first fit the existing data.
We evaluate the accuracy of the models under comparison by calculating the relative prediction error at each time slice:
\[
\|\ten{X}_t - [\![\mat{A}^{(1)},\dots,\mat{A}^{(N)},\hat{\mat{a}}^{(N+1)}_{t}]\!] - \ten{S}_t\|_{\rm F} /\|\ten{X}_t\|_{\rm F}.
\]

Fig.~\ref{fig:network factorization} compares different methods on the full dataset with a ``burn-in" time of 10 frames, after which the error patterns are stable. Our algorithm significantly outperforms OLSTEC and Online-SGD in factoring the whole data set.
\begin{figure}[t]
\begin{center}
\includegraphics[width=.47\textwidth]{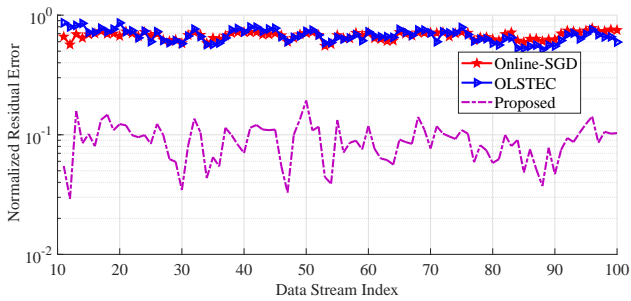}
\caption{Factorization error of network traffic from complete samples.}
\label{fig:network factorization}
\end{center}
\end{figure}
Then we remove $50\%$ of the entries from the the Abilene tensor and attempt to reconstruct the whole network traffic. Our results are shown in Fig.~\ref{fig:network completion}. Again we use a ``burn-in" time of 10 frames. 

\begin{figure}
\begin{center}
\includegraphics[width=.47\textwidth]{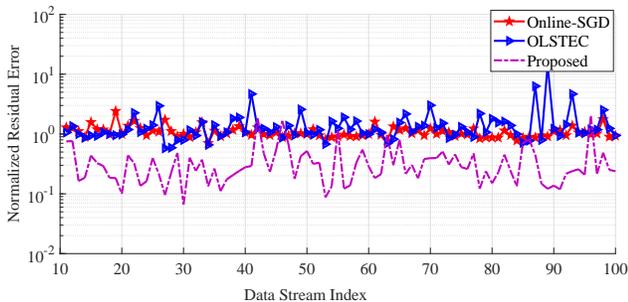}
\caption{Reconstruction error of network traffic with $50\%$ of data missing.}
\label{fig:network completion}
\end{center}
\end{figure}

\section{Conclusion}

We have proposed a Bayesian formulation to the problem of robust streaming tensor completion and factorization. The main advantages of our algorithm are {\bf automatic rank determination} and {\bf robustness} to outliers. We have demonstrated the benefits of robustness on MRI and network flow data.  Due to the automatic rank determination and the robustness to outliers, our algorithm has achieved higher accuracy than existing approaches on all tested streaming tensor examples.  

\section{Acknowledgements}
We thank the anonymous referees for their helpful comments. A special thanks to Chunfeng Cui for many suggestions to improve this manuscript. This work was partially supported by  NSF CCF Award No. 1817037.

{\small
\bibliographystyle{IEEEtran}
\bibliography{egbib}
}

\end{document}